\documentclass[letterpaper]{article} 
\usepackage{aaai2027}  
\usepackage[hyphens]{url}  
\usepackage{graphicx} 
\usepackage{natbib}  
\usepackage{caption} 
\usepackage{algorithm}
\usepackage{algorithmic}
\usepackage{graphicx}
\usepackage{booktabs}
\usepackage{multirow}
\usepackage{amsmath}
\usepackage{color}

\usepackage{newfloat}
\usepackage{listings}
\DeclareCaptionStyle{ruled}{labelfont=normalfont,labelsep=colon,strut=off} 
\floatstyle{ruled}
\newfloat{listing}{tb}{lst}{}
\floatname{listing}{Listing}

\usepackage{booktabs}

\title{Dense Point-to-Mask Optimization with Reinforced Point Selection for \\ Crowd Instance Segmentation}
\author {
    Hongru Chen\textsuperscript{\rm 1},
    Jiyang Huang\textsuperscript{\rm 1},
    Jia Wan\textsuperscript{\rm 1}\corresponding,
    Antoni B.Chan\textsuperscript{\rm 2}
}
\affiliations{
    \textsuperscript{\rm 1} Harbin Institute of Technology, Shenzhen\\
    \textsuperscript{\rm 2} City University of Hong Kong \\
     {\{24S151003\}@stu.hit.edu.cn}\\
    {\{jiyanghuang0127, jiawan1998\}@gmail.com}\\
    {\{abchan\}@cityu.edu.hk}
}

\begin{document}

\maketitle

\begin{abstract}
Crowd analysis is a crucial task with a wide range of applications, including surveillance and transportation. 
Currently, point labels are common in crowd datasets, while region labels (e.g., boxes) are rare and inaccurate. 
Although segmentation foundation models offer a potential route from points to masks, directly prompting them in dense crowds frequently produces merged, overlapping, or identity-ambiguous instances.
To this end, we first propose Dense Point-to-Mask Optimization (DPMO), 
a geometry-constrained mask generation method that converts crowd point annotations into instance masks. 
DPMO couples the segmentation prior of the Segment Anything Model with a Nearest-Neighbor Exclusive Circle (NNEC) constraint derived solely from point geometry. The constraint establishes an instance region for every annotation, suppressing cross-instance leakage and preserving point-to-mask identity under severe crowding. 
Combined with lightweight manual correction, DPMO enables us to augment four established crowd benchmarks with instance-mask annotations.
Then, we propose a Reinforced Point Selection (RPS) for dense crowd instance segmentation
which learns to select the candidate that provides the most effective instance cue. Candidate selection is optimized with modified Group Relative Policy Optimization (GRPO) using mask-derived rewards, directly aligning point selection with downstream instance segmentation quality. 
Finally, we use the generated masks to define instance-aware supervision for crowd counting, allowing conventional counting models to learn object extent and separation in addition to density.
Through extensive experiments, we achieve state-of-the-art crowd instance segmentation and counting performance on ShanghaiTech, UCF-QNRF, JHU-CROWD++, and NWPU-Crowd datasets. 
\end{abstract}
\vspace{-0.4cm}
\section{Introduction}
\label{sec:intro}
Crowd analysis is increasingly attracting attention due to its wide application in social governance and security maintenance. 
Masks obtained through instance segmentation constitute a significant asset to the field of crowd analysis.
Traditional crowd analysis tasks, such as crowd counting, mainly obtain crowd distribution through individual point coordinates~\cite{liu2019point} and overall density maps~\cite{Kang2017BeyondCC, Li2021ApproachesOC}. 
However, point coordinates and density maps provide only crowd quantity and individual location, failing to capture fine-grained details such as individual shapes and inter-person variations.
Furthermore, since the point coordinates and the density map lack correspondence, conventional pixel aggregation methods are inadequate for establishing reliable associations between them.
Therefore, we propose crowd instance segmentation to obtain more detailed crowd information by segmenting individuals within the crowd, compensating for the shortcomings of the original point coordinates and density maps, and achieving more accurate crowd analysis.

Recently, large foundation models have emerged and are being applied to various visual tasks. The Segment Anything Model (SAM)~\cite{kirillov2023segment} has emerged as a powerful zero-shot segmenter, providing a new method for mask generation. However, in situations with extremely dense or heavily occluded areas, the head can become small and significantly obscured. In such cases, directly using SAM~\cite{kirillov2023segment} may hinder the accurate separation of individuals from the background, ultimately impacting segmentation accuracy.

In this work, we establish \emph{crowd instance segmentation} as a systematic benchmark task, where the objective is to predict an individual mask for every person in a crowd. We augment ShanghaiTech, UCF-QNRF, JHU-CROWD++, and NWPU-Crowd with human-verified instance-mask annotations. To make annotation feasible, we propose Dense Point-to-Mask Optimization (DPMO), a human-in-the-loop pipeline that converts existing point labels into editable mask proposals. DPMO combines the segmentation prior of SAM with a Nearest-Neighbor Exclusive Circle (NNEC) constraint derived from local point geometry. The constraint defines an instance-specific feasible region and reduces leakage into neighboring people. Human annotators then inspect and correct the proposals to ensure annotation reliability.

We further propose Reinforced Point Selection (RPS) for crowd instance segmentation. In dense scenes, small changes in a predicted point can produce substantially different masks, and the point closest to the annotation is not necessarily the best segmentation prompt. RPS samples multiple candidate points and learns to select the candidate that produces the highest-quality instance mask. It is trained using a modified Group Relative Policy Optimization objective with mask-based rewards, directly aligning point selection with downstream segmentation performance.
Finally, we investigate the value of instance masks for crowd counting. Unlike conventional point-generated density targets, masks encode object extent and separation. We introduce mask-supervised objectives that can be incorporated into different counting architectures. Experiments on the four benchmarks demonstrate state-of-the-art crowd instance-segmentation performance and consistent counting improvements, highlighting the broader value of instance-level supervision for crowd analysis.

\begin{figure*}[t]
  \centering
  \setlength{\abovecaptionskip}{-0.005cm} 
  \setlength{\belowcaptionskip}{-0.2cm} 
  \includegraphics[width=\textwidth]{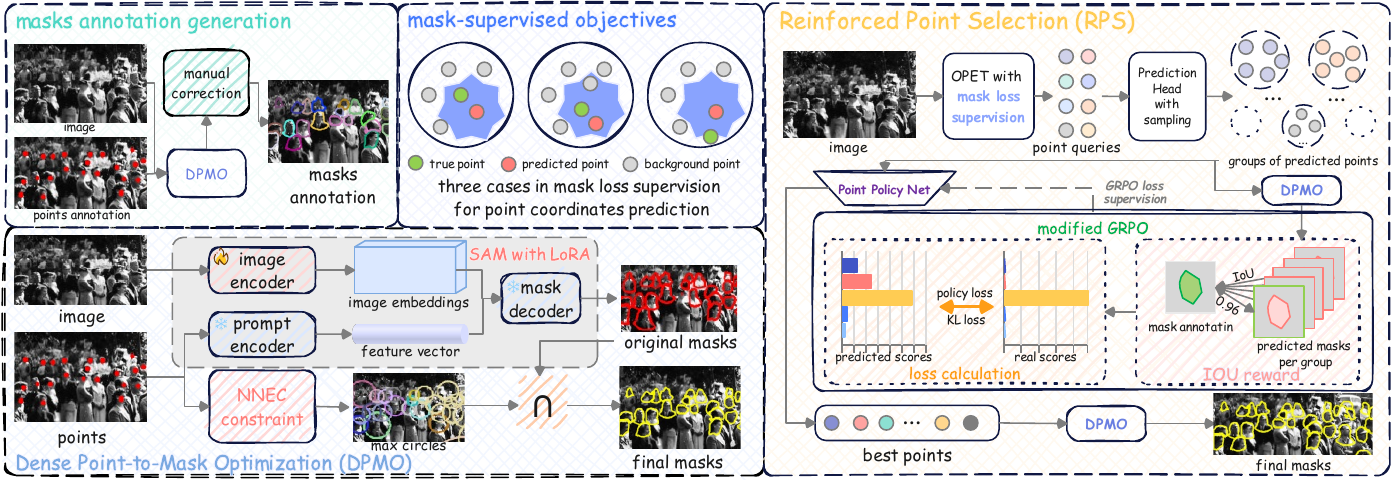}
  \caption{Pipeline of our proposed DPMO-RPS framework for crowd instance segmentation, including DPMO segmentation paradigm using NNEC constraint and RPS network trained with modified GRPO. Supervised by manually modified mask annotations, we train the OPET Net to obtain more accurate predicted points and perform sampling. Then, we train the Point Policy Network with modified GRPO to select the best point from each group as a final point prompt to assist SAM in segmentation in DPMO. Finally, we can achieve effective crowd instance segmentation. } 
  \label{fig:overview}
\end{figure*}

The contributions of this paper are summarized as:
\begin{itemize}
  \item We establish \emph{crowd instance segmentation} as a systematic benchmark task that aims to predict an individual mask for every person in a crowded scene. 
  This formulation extends conventional point-based crowd analysis to instance-level spatial understanding, enabling explicit point-to-instance correspondence and providing richer supervision for crowd counting.
  \item 
  We construct human-verified instance-mask annotations for four widely used crowd datasets. To make large-scale annotation feasible, we propose Dense Point-to-Mask Optimization (DPMO), a human-in-the-loop annotation pipeline that combines the segmentation prior of SAM with a Nearest-Neighbor Exclusive Circle (NNEC) constraint. DPMO converts existing point labels into instance-specific mask proposals, which are subsequently inspected and corrected by human annotators to ensure reliability.
  \item We propose Reinforced Point Selection (RPS) framework for crowd instance segmentation. RPS samples multiple candidate point predictions and learns to select the candidate that produces the highest-quality instance mask. A mask-based reward and a modified Group Relative Policy Optimization objective directly align the point-selection policy with downstream segmentation performance.
  \item Extensive experiments on ShanghaiTech, UCF-QNRF, JHU-CROWD++, and NWPU-Crowd demonstrate state-of-the-art crowd instance-segmentation performance. We further develop mask-supervised objectives for crowd counting and obtain consistent improvements across multiple counting architectures, demonstrating the value of instance-mask supervision beyond the proposed segmentation task. 
\end{itemize}

\vspace{-0.4cm}
\section{Related Works}

\textbf{Instance Segmentation}
\quad Instance segmentation \cite{Ren2015FasterRT, li2016iterative, bolya2019yolact, wang2020solov2} is receiving increasing attention due to its wide range of potential applications in areas such as surveillance, traffic management, and commerce. In extremely dense scenes, these models typically require lower thresholds and smaller subdivisions to perform instance segmentation. The emergence of large-scale visual models provides new methods for instance detection and segmentation. 

The Contrastive Language-Image Pre-training model (CLIP)~\cite{radford2021learning,jia2021scaling} connects vision and language via contrastive image-text pre-training, which can be applied to any image classification benchmark by providing the names of the visual categories.
The Segment Anything Model (SAM)~\cite{kirillov2023segment}, with its powerful zero-shot segmentation capabilities by introducing promptable concept, provides a novel approach and strategy for instance segmentation. \cite{Wei2023SemanticawareSF, zhao2023fast, Chen2023RSPrompterLT} have improved the performance of instance segmentation by optimizing the SAM~\cite{kirillov2023segment} model. 
The EVA~\cite{fang2023eva} serves as a powerful vision encoder for instance segmentation by pioneering a pre-training task that reconstructs features from masked images.
There are also many models proposed for instance segmentation of individuals. \cite{Zhang2023PersonalizeSA, cai2024crowd} identified and segmented people from many targets.

However, due to problems such as mutual occlusion between individuals or the small size of human heads making them difficult to distinguish from the background in extremely complex scenarios such as dense crowds, the segmentation performance of directly using the above segmentation models to segment crowds on crowd datasets is not ideal.

\noindent \textbf{Crowd Counting}
\quad Traditional models can be divided into two categories based on whether they generate crowd instance-level information (i.e., bounding boxes or points): density map-based prediction networks and point coordinate-based prediction networks. 

Density map-based methods predict crowd count by generating a probability density map, where integrating pixel values yields the total number of people. \cite{zhang2015cross, zhang2016single, sindagi2017generating, Zhao2024ADC} have improved the prediction accuracy of density maps utilizing multi-layer neural networks and contextual information. Given that traditional density maps fail to distinguish individual heads in crowded areas due to occlusion, \cite{Cheng2021two-stage, Khan2022CrowdCA} proposed algorithms that identify potential head locations and then perform accurate counting based on those locations. With the help of head masks, \cite{Jiang2018MaskAwareNF, Yao2020MaskGG, Duan2021MaskbasedGA, Ren2022EnhancementOL} generated the enhanced density map, which can effectively solve the problem of low recall of head detection.  After the advent of the transformer, \cite{Chen2022CountingVD, Wang2025GlobalVL} attempted to apply this advanced technology to crowd counting tasks. \cite{jiang2023clip, Liang2023CrowdCLIPUC, ma2024clip} utilized the cross-modal alignment capability of the CLIP model~\cite{radford2021learning} for crowd counting using text constraints. 

Point coordinates-based networks directly predict head bounding boxes~\cite{liu2019point} or head coordinates~\cite{Lian2019Detection}, deriving crowd count by counting these predictions. Researchers, such as~\cite{song2021rethinking, Sam2021Locate}, have completely abandoned traditional density maps and directly estimated the crowd size by statistically analyzing the point coordinates. \cite{liu2023point} introduced a point-query quad-tree mechanism, which allows the model to proactively generate sparse point queries and adaptively focus on regions where targets might exist. \cite{Chen2024ImprovingPC, Ranasinghe2024CrowdDV} achieved more accurate crowd counting by improving the accuracy of model localization. While such networks can locate individuals, they lack fine-grained information like the shape and size of individuals. 
Therefore, to achieve crowd instance segmentation, it is necessary to combine a segmentation model to obtain more granular detailed information.

Conventional crowd counting techniques struggle to predict the shape and scale of individuals within a crowd. Additionally, existing segmentation methods tend to perform poorly in dense scenes. To address these limitations, we propose a Dense Point-to-Mask Optimization with Reinforced Point Selection (DPMO-RPS) framework that generates non-overlapping masks for individuals, resulting in effective crowd instance segmentation.

\vspace{-0.3cm}
\section{Method}

We propose Dense Point-to-Mask Optimization with Reinforced Point Selection (DPMO-RPS), a unified framework for predicting an individual mask for each person in a crowded scene. 
As illustrated in Fig.~\ref{fig:overview}, the framework comprises two main components. 
First, DPMO combines an optimized SAM segmentation module with the proposed Nearest-Neighbor Exclusive Circle (NNEC) constraint to convert point prompts into mutually exclusive instance masks while preserving their one-to-one correspondence with individual points. DPMO also serves as an efficient annotation tool: it generates initial mask proposals from existing point labels, which are subsequently corrected by human annotators to ensure reliability. 
Second, RPS samples multiple candidate points for each instance and learns to select the point that produces the highest-quality mask when used as a DPMO prompt. RPS is optimized using a modified GRPO objective~\cite{shao2024deepseekmath} with segmentation-quality rewards. The complete DPMO-RPS framework is then trained end-to-end to jointly optimize point selection and instance-mask prediction for dense crowds.

\begin{table*}[tbp]
\centering
\vspace*{-3mm}
\setlength{\abovecaptionskip}{-0.01cm}
\setlength{\belowcaptionskip}{-0.5cm}
\resizebox{\textwidth}{!}{%
\tiny
\begin{tabular}{llclclclclc}
\hline
dataset &  & model              &  & IoU $\uparrow$  &  & Precision $\uparrow$       &  & Recall $\uparrow$          &  & F1 $\uparrow$              \\ \hline
\multirow{4}{*}{SHA~\cite{zhang2016single}}        &  & SAM \cite{kirillov2023segment}+Classification &  & 0.3987          &  & 0.3630          &  & 0.3498          &  & 0.3563          \\
        &  & FastSAM~\cite{zhao2023fast}             &  & 0.3624          &  & 0.3034          &  & 0.2748          &  & 0.2884          \\
     &  & CrowdSAM~\cite{cai2024crowd}           &  & 0.3530          &  & 0.2997          &  & 0.1422          &  & 0.1928          \\
        &  & \textbf{Ours}      &  & \textbf{0.5431} &  & \textbf{0.5495} &  & \textbf{0.5572} &  & \textbf{0.5533} \\ \hline
\multirow{4}{*}{SHB~\cite{zhang2016single}} &  & SAM~\cite{kirillov2023segment}+Classification &  & 0.3776          &  & 0.3223          &  & 0.2606          &  & 0.2882           \\
                     &  & FastSAM~\cite{zhao2023fast}            &  & 0.4269          &  & 0.4220          &  & 0.4397            &  & 0.4307        \\
                     &  & CrowdSAM~\cite{cai2024crowd}           &  & 0.2116          &  & 0.0901          &  & 0.0959            &  & 0.0929         \\
                     &  & \textbf{Ours}      &  & \textbf{0.6201} &  & \textbf{0.6412} &  & \textbf{0.6384} &  & \textbf{0.6398} \\ \hline
\multirow{4}{*}{UCF~\cite{Idrees2018UCF-QNRF}}        &  & SAM~\cite{kirillov2023segment}+Classification &  & 0.3102          &  & 0.1288          &  & 0.2012          &  & 0.1571          \\
        &  & FastSAM~\cite{zhao2023fast}           &  & 0.3708          &  & 0.3061          &  & 0.1891          &  & 0.2338           \\
     &  & CrowdSAM~\cite{cai2024crowd}           &  & 0.3283          &  & 0.2648          &  & 0.0763          &  & 0.1184          \\
        &  & \textbf{Ours}      &  & \textbf{0.4769} &  & \textbf{0.4426} &  & \textbf{0.4396} &  & \textbf{0.4411} \\ \hline
\multirow{4}{*}{JHU~\cite{sindagi2020jhu}} &  & SAM~\cite{kirillov2023segment}+Classification &  & 0.2116          &  & 0.1892          &  & 0.1405                &  & 0.1613                \\
                     &  & FastSAM~\cite{zhao2023fast}            &  & 0.3766          &  & 0.2211          &  & 0.2205            &  & 0.2208                \\
                     &  & CrowdSAM~\cite{cai2024crowd}           &  & 0.3091          &  & 0.2154          &  & 0.1153            &  & 0.1502                \\
                     &  & \textbf{Ours}      &  & \textbf{0.5218}       &  & \textbf{0.4286}       &  & \textbf{0.4068}       &  & \textbf{0.4174}       \\ \hline
\multirow{4}{*}{NWPU~\cite{Wang2021NWPU_Crowd}}        &  & SAM~\cite{kirillov2023segment}+Classification &  & 0.3599           &  & 0.2777          &  & 0.1983         &  & 0.2314      \\
        &  & FastSAM~\cite{zhao2023fast}           &  & 0.2694          &  & 0.1384          &  & 0.1094          &  & 0.1222           \\
    &  & CrowdSAM~\cite{cai2024crowd}           &  & 0.2719          &  & 0.1830          &  & 0.0811          &  & 0.1124          \\
 &  & \textbf{Ours} &  & \multicolumn{1}{l}{\textbf{0.5023}} &  &\multicolumn{1}{c}{\textbf{0.4966}} &  & \multicolumn{1}{l}{\textbf{0.4897}} &  & \multicolumn{1}{l}{\textbf{0.4931}} \\ \hline
\end{tabular}%
}
\caption{Segmentation performance comparison between our model and other models on crowd datasets. } 
\label{tab:Contrast with state-of-the-arts}
\end{table*}

\vspace{-0.3cm}
\subsection{Dense Point-to-Mask Optimization (DPMO)}
We propose a DPMO framework for instance segmentation in crowds, using optimized SAM~\cite{kirillov2023segment} with NNEC constraints.
SAM~\cite{kirillov2023segment} cannot accurately segment individuals in a crowd image because of the small size and occlusion in dense areas. Furthermore, the pixels in these areas are difficult to distinguish as belonging to the foreground or background.
Therefore, in dense scenes of the image, the segmentation mask 
may be overlapping and incomplete.
To this end, we propose NNEC constraint to optimize the scale of segmentation masks in dense regions.

The NNEC constraint aims to find the maximum possible range of the segmentation mask for each point annotation. Since each segmentation mask corresponds uniquely to a specific person, it should contain only one annotation point. 
To better cover head regions, circles are allowed to overlap with their radii dynamically adjusted according to distances to neighboring points.
The formula of NNEC constraint is as follows:
\begin{equation}
    \small
    \forall i, \forall j \in \mathcal{N}_K(i), \quad r_i \leq \| \mathbf{x}_i - \mathbf{x}_j \|_2,
    \label{eq:NNEC algorithm}
\end{equation}
where $x_i$ and $x_j$ respectively represent different predicted points in the set, $r_i$ stands for the maximum possible range of the mask for point $x_i$, 
and $\mathcal{N}_K(i)$ is a set composed of the K other points that are closest to this point, which is defined as follows: 
\begin{equation}
    \small
    \mathcal{N}_K(i) = \{ j \mid j \neq i, \; \| \mathbf{x}_i - \mathbf{x}_j \|_2 \leq d_i^{(K)} \},
\end{equation}
where $d_i^{(K)}$ represents the Euclidean distance from point $x_i$ to its K-th nearest neighbor.
Geometrically, this constraint serves to constrain the masks, preventing them from encompassing multiple points, while also aiding in the elimination of points that were inadequately segmented into masks.
An empty intersection between segmentation and the constrained circle indicates that the initial segmentation 
either failed to yield a mask or that the resulting mask does not fall within the expected range. In such instances, the largest circle is adopted as the final mask. Therefore, limiting the range of the circle radius in NNEC constraint is crucial, as different radius ranges directly affect the accuracy of final segmentation masks.

With this DPMO framework, we obtain robust segmentation masks from original point annotations in crowd counting datasets. After manually correcting these masks, as shown in Fig.~\ref{fig:overview}, we can obtain reliable mask annotations that correspond one-to-one with the original point annotations but contain more fine-grained information about the individuals.

\vspace{-0.3cm}
\subsection{Reinforced Point Selection (RPS)}
The quality of a prompt-based mask is highly sensitive to the position of its point prompt, particularly in dense crowds where adjacent instances have similar appearances and substantial overlap. A point with a small localization error may still fall on an occluded or ambiguous region and consequently produce an inaccurate mask. To address this mismatch between localization accuracy and segmentation quality, we propose Reinforced Point Selection (RPS). Built upon DPMO, RPS generates multiple point candidates for each detected person and learns to select the candidate that produces the most accurate instance mask. As illustrated in Fig.~\ref{fig:overview}, RPS comprises two components: Optimized PET (OPET) for candidate generation and a Point Policy Network (PPNet) for candidate selection.

\vspace{-0.2cm}
\paragraph{Mask-Supervised Point Prediction}
We construct OPET by extending the PET point-localization model~\cite{liu2023point}. The original PET supervises point prediction through bipartite matching based primarily on classification confidence and coordinate distance. However, coordinate proximity alone does not guarantee that a predicted point lies within the visible region of its corresponding person. We introduce a mask-supervised loss that encourages each matched prediction to remain inside its corresponding ground-truth instance mask. This supervision reduces predictions located on the background or neighboring instances, making the predicted points more suitable as segmentation prompts.

Because a single deterministic prediction may still be suboptimal for segmentation, we construct a candidate group around each initial OPET prediction. For the $i$-th detected instance, OPET produces an initial point $\mathbf{p}_{i,1}$. We then sample $K-1$ additional points from a Gaussian distribution centered at the initial prediction:
\begin{equation}
    \small
    \mathbf{p}_{i,c}
    =
    \mathbf{p}_{i,1}+\boldsymbol{\delta}_{i,c},
    \boldsymbol{\delta}_{i,c}
    \sim
    \mathcal{N}(\mathbf{0},\boldsymbol{\Sigma}_{i}),
    c=2,\ldots,K.
\end{equation}

We set $K=5$ in our experiments. The initial prediction and four sampled points form the candidate set $\mathcal{P}_{i}=\{\mathbf{p}_{i,c}\}_{c=1}^{K}$. This sampling strategy explores nearby prompt positions while preserving the identity of the predicted instance.

\vspace{-0.2cm}
\paragraph{Segmentation-Based Reward}
Each candidate point $\mathbf{p}_{i,c}$ is passed to DPMO to produce a candidate mask $M_{i,c}$. Its segmentation score is calculated using the intersection over union between the candidate mask and the corresponding ground-truth mask $M_{i}^{\mathrm{gt}}$:
\begin{equation}
    \small
    s_{i,c}
    =
    \operatorname{IoU}
    \left(
        M_{i,c},M_{i}^{\mathrm{gt}}
    \right).
\end{equation}

The scores are normalized within each candidate group to obtain group-relative advantages:
\begin{equation}
    \resizebox{0.43\textwidth}{!}{
    $ A_{i,c}
    =
    \frac{s_{i,c}-\mu_i}{\sigma_i+\delta},
    \mu_i
    =
    \frac{1}{K}
    \sum_{c=1}^{K}s_{i,c},
    \sigma_i
    =
    \sqrt{
        \frac{1}{K}
        \sum_{c=1}^{K}
        \left(s_{i,c}-\mu_i\right)^2
    },$
    }
\end{equation}
where $\delta$ is a small constant introduced for numerical stability. The advantage $A_{i,c}$ indicates whether a candidate performs better or worse than the other candidates generated for the same instance. This group-relative formulation eliminates the need to train a separate reward model.

\vspace{-0.2cm}
\paragraph{Point Policy Optimization}
PPNet takes the image features, instance representation, and candidate coordinates as the state $x_i$. It then predicts a categorical probability distribution over the $K$ candidate points:
\begin{equation}
    \small
    \pi_\theta(a_{i,c}\mid x_i)
    =
    P\left(
        \mathbf{p}_{i,c}
        \mid
        x_i,\mathcal{P}_i
    \right),
\end{equation}
where $a_{i,c}$ denotes the action of selecting the $c$-th candidate from $\mathcal{P}_i$. We optimize PPNet using a modified Group Relative Policy Optimization objective~\cite{shao2024deepseekmath}. The probability ratio between the current policy and the previous policy is defined as
\begin{equation}
    r_{i,c}(\theta)
    =
    \frac{
        \pi_\theta(a_{i,c}\mid x_i)
    }{
        \pi_{\theta_{\mathrm{old}}}(a_{i,c}\mid x_i)
    },
\end{equation}

The complete RPS objective is defined as
\begin{equation}
    \mathcal{L}_{\mathrm{RPS}}
    =
    \mathcal{L}_{\mathrm{policy}}
    -
    \beta\mathcal{L}_{\mathrm{KL}},
\end{equation}
where $\beta$ controls the strength of the KL-divergence regularization.
For a mini-batch containing $G$ candidate groups, the clipped policy loss is
\begin{equation}
\resizebox{0.43\textwidth}{!}{%
\begin{minipage}{0.6\textwidth}
$\mathcal{L}_{\mathrm{policy}} = \frac{1}{GK} \sum_{i=1}^{G}\sum_{c=1}^{K} \min \big[ r_{i,c}(\theta)A_{i,c}, \operatorname{clip}(r_{i,c}(\theta), 1-\epsilon, 1+\epsilon) A_{i,c} \big]$
\end{minipage}%
}
\end{equation}
where $\epsilon$ controls the range of each policy update. To prevent the current policy from deviating excessively from the previous policy, we introduce a KL-divergence regularizer:
\begin{equation}
    \resizebox{0.42\textwidth}{!}{
    \small
    $ \mathcal{L}_{\mathrm{KL}}
    =
    \frac{1}{G}
    \sum_{i=1}^{G}
    D_{\mathrm{KL}}
    \left(
        \pi_{\theta_{\mathrm{old}}}
        (\cdot\mid x_i)
        \,\Vert\,
        \pi_\theta
        (\cdot\mid x_i)
    \right),$
    }
\end{equation}

During inference, OPET generates an initial point and $K-1$ sampled candidates for each detected person. PPNet evaluates the complete candidate group, and the candidate with the highest selection probability is chosen as the final point prompt:
\begin{equation}
    \small
    c_i^{*}
    =
    \underset{c\in\{1,\ldots,K\}}{\arg\max}
    \ \pi_\theta(a_{i,c}\mid x_i),
    \quad
    \mathbf{p}_i^{*}
    =
    \mathbf{p}_{i,c_i^{*}}.
\end{equation}
The selected point $\mathbf{p}_i^{*}$ is subsequently passed to DPMO to generate the final instance mask. By optimizing point selection according to segmentation quality rather than coordinate distance alone, RPS produces point prompts that are more robust to localization deviations and more suitable for instance segmentation in dense crowds.

\begin{table*}[t]
\centering 
\vspace*{-3mm}
\setlength{\abovecaptionskip}{-0.01cm}
\setlength{\belowcaptionskip}{-0.3cm}
\resizebox{\linewidth}{!}{%
\tiny
\begin{tabular}{cccccccccccccc}
\hline
\multirow{2}{*}{model} & 
  \multirow{2}{*}{Loss Function} & 
  \multicolumn{2}{c}{ShanghaiTech A} & 
  \multicolumn{2}{c}{UCF-QNRF} & 
  \multicolumn{2}{c}{NWPU-Crowd} \\ \cline{3-8} 
                     &              &  MAE $\downarrow$         &  RMSE $\downarrow$  &  MAE $\downarrow$       &  RMSE $\downarrow$      &  MAE $\downarrow$    & RMSE $\downarrow$   \\ \hline
{CSRNet~\cite{li2018csrnet}} & - & 68.2 & 115.0 & - & - & 121.3 & 387.8 \\
{BL+~\cite{Ma2019BayesianLoss}} & - & 62.8 & 101.8 & 88.7 & 154.8 & 105.4 & 454.2 \\
{DMCount~\cite{wang2020distribution}} & - & 59.7 & 95.7 & 85.6 & 148.3  & 88.4 & 388.6 \\
{P2PNet~\cite{song2021rethinking}} & - & 52.74 &85.06 &85.32 &154.5 &77.44 &362.0 \\ 
{GL~\cite{Wan2021GLloss}} & - & 61.3 & 95.4 & 84.3 & 147.5 & 79.3 & 346.1 \\
{CLTR~\cite{liang2022end}} & - & 56.9 & 95.2 & 85.8 & 141.3 & 74.4 & 333.8 \\ \hline
\multirow{2}{*}{PET~\cite{liu2023point}} & original loss &  49.34     &  78.77 &  79.53   &  144.32   & 74.4   &  328.5 \\
                     &{Ours}  &{48.21} &{77.35} &{77.98}  &{142.18}  & {73.34}   &{307.70}   \\ \hline
\multirow{2}{*}{CLIP-EBC~\cite{ma2024clip}} 
                     &  original loss &  54.0 & {83.2} &  80.5 &  136.6 &  75.8 &  367.4\\
                     &  {Ours}          &  {52.6} &  85.1 &  {77.1} 
                     & {135.7} &  {74.26} &  {345.6} \\ \hline
\multirow{2}{*}{ZIP~\cite{Ma2025ZIPSC}} &  original loss &  \underline{47.81}   &  \underline{75.04}  &  \underline{69.46}  &  \underline{121.88} &  \underline{60.1} &  \underline{301.2} \\
                     &  {Ours} &  \textbf{46.95}  &  \textbf{73.83}  &  \textbf{68.55} &  \textbf{120.68}   &  \textbf{60.02} &  \textbf{278.64} \\ \hline
\end{tabular}%
}
\caption{Results of crowd counting comparison using the proposed mask loss function supervised by mask annotations.}
\label{tab:loss function results}
\end{table*}

\begin{figure*}[t]
  \centering
  \setlength{\abovecaptionskip}{-0.005cm} 
  \setlength{\belowcaptionskip}{-0.5cm} 
  \includegraphics[width=\textwidth]{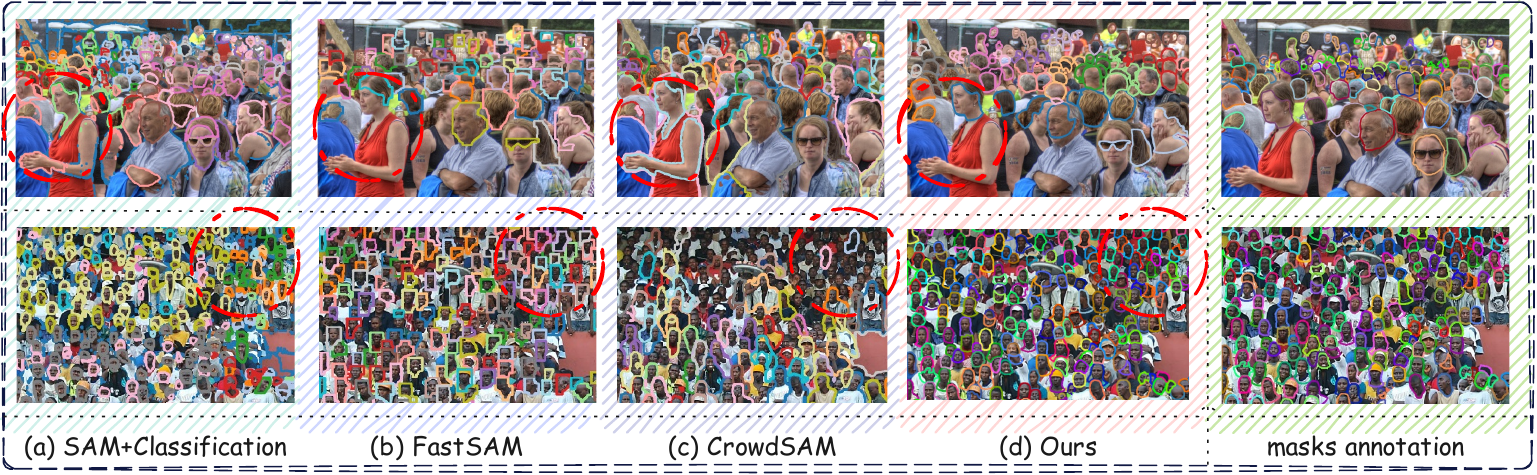}
  \caption{Segmentation results contrast with different models, including the SAM-optimized model, FastSAM, CrowdSAM, and our DPMO-RPS model. The robust mask annotations are in the rightmost column. }
  \label{fig:Results Contrast}
\end{figure*}

\vspace{-0.3cm}
\subsection{Mask Supervision for Crowd Counting}
\label{subsec:loss function}

Following the proposed DPMO-RPS algorithm, we obtain reliable instance-level mask annotations from the original point annotations with only minimal manual corrections. Since each segmentation mask corresponds uniquely to one annotated head point, the ambiguity between point annotations and density maps is eliminated. Compared with point annotations, masks provide richer spatial information of each individual, while offering more precise localization than density maps. Based on these properties, we design two simple yet effective mask-supervised loss functions for two representative categories of crowd counting networks.

\vspace{-0.15cm}
\subsubsection{Loss Function for Density Map Prediction}

Density map-based crowd counting networks estimate a continuous density map, where the value of each pixel represents the local crowd density. Ideally, the integral of the predicted density within each individual's mask should equal one, while the density outside all masks should be zero.

Specifically, let $\mathcal{M}_i$ denote the $i$-th segmentation mask, $\mathcal{B}$ denote the background region, and $D(p)$ denote the predicted density value at pixel $p$. We define the proposed mask supervision loss as
\begin{equation}
    \resizebox{0.40\textwidth}{!}{
    $ \mathcal{L}_{\mathrm{mask}}
    =
    \frac{1}{N}
    \sum_{i=1}^{N}
    \left(
    \sum_{p\in\mathcal{M}_i}D(p)-1
    \right)^2
    +
    \lambda
    \left(
    \sum_{p\in\mathcal{B}}D(p)
    \right)^2, $
    }
\end{equation}
where $N$ denotes the number of masks in the image and $\lambda$ is a balancing coefficient for the background constraint. The first term enforces each individual mask to contain exactly one person by constraining the density integral to one, while the second term suppresses false responses in the background. Consequently, the predicted density distribution gradually converges to the spatial distribution defined by the segmentation masks during training.

\vspace{-1.5mm}
\subsubsection{Loss Function for Point Coordinate Prediction}
Point-based crowd counting methods directly predict head locations. Their training relies on establishing correspondences between predicted points and ground-truth annotations. Unlike conventional methods that employ Hungarian matching, we exploit the instance-level mask annotations to perform geometry-aware point matching.

Since each mask uniquely corresponds to one ground-truth point, a valid prediction should fall inside its associated mask region. 
Accordingly, the matching strategy consists of three cases, as illustrated in Fig.~\ref{fig:overview}:
(1) if exactly one predicted point lies inside a mask, it is directly matched with the corresponding ground-truth point.
(2) if multiple predicted points fall inside the same mask, the one closest to the ground-truth point is selected.
(3) if no predicted point lies inside the mask, the nearest prediction outside the mask is assigned.
Based on this strategy, point assignment is formulated as a constrained optimization problem,
\begin{equation}
    \resizebox{0.40\textwidth}{!}{
    $ \min_{\mathbf{X}}
    \sum_{i=1}^{m}\sum_{j=1}^{n}
    c_{ij}x_{ij},
    \quad c_{ij}=
    \begin{cases}
    d_{ij}, & p_j\in\mathcal{M}_i,\\
    d_{ij}+\gamma, & p_j\notin\mathcal{M}_i,
    \end{cases} $
    \label{eq:matching_cost}
    }
\end{equation}
where $m$ and $n$ denote the numbers of ground-truth and predicted points, respectively, $x_{ij}\in\{0,1\}$ indicates whether prediction $j$ is assigned to ground-truth point $i$, 
$d_{ij}$ denotes the Euclidean distance between the predicted point and the ground-truth point, and $\gamma$ is a penalty applied to predictions located outside the corresponding mask. This formulation encourages predictions to remain within the correct instance region while minimizing localization error.

We further apply this mask-supervised matching loss to train the OPET network. As demonstrated in ablation studies, the proposed supervision consistently achieves better performance than conventional point-annotation supervision.

\vspace{-0.3cm}
\section{Experiments}
\quad This section presents the experimental settings, including the evaluation protocol, data processing, and training details. We compare the proposed DPMO-RPS framework with state-of-the-art methods for crowd instance segmentation and evaluate the effectiveness of the proposed mask-supervised loss function for crowd counting. Finally, ablation studies are conducted to investigate the effects of different annotations, point prediction models, and key hyperparameters.

\vspace{-0.3cm}
\subsection{Evaluation Criteria}
\label{subsec:eval}

We evaluate crowd instance segmentation from both mask quality and overall segmentation performance using IoU, Precision, Recall, and F1-score.

First, the quality of segmentation masks is measured by Intersection over Union (IoU). For each image, an IoU matrix is computed between the predicted masks and the ground-truth mask annotations. The Hungarian algorithm is then applied to the IoU matrix to obtain the optimal one-to-one matching between predicted and ground-truth masks. The final IoU is calculated as the average IoU of all matched mask pairs over the entire dataset.

Second, the overall segmentation performance is evaluated using the Precision, Recall, and F1-score derived from the confusion matrix.
The specific formula is as follows:
\begin{tiny}
\begin{align} 
  \text{Precision} &= \frac{\text{TP}}{\text{TP} + \text{FP}}, \quad 
  \text{Recall} &= \frac{\text{TP}}{\text{TP} + \text{FN}},\quad
  \text{F1} &= \frac{2\times\left ( \text{Precision}\times \text{Recall}\right ) }{\text{Precision} + \text{Recall}},
  \label{eq:important}
\end{align}
\end{tiny}
where TP, FP, and FN denote the numbers of true positives, false positives, and false negatives, respectively.

\begin{table}[t]
\centering
\setlength{\abovecaptionskip}{-0.01cm}
\setlength{\belowcaptionskip}{-0.2cm}
\resizebox{\columnwidth}{!}{%
\begin{tabular}{ccccc}
\hline
datasets & time (hours) & correction ratio (\%) & correction consistency (IoU) & quality gap (IoU) \\ \hline
SHA      & 8.5               & 22.5             & 0.9823           & 0.9588            \\
SHB      & 12.5              & 18.0             & 0.9934           & 0.9762            \\
UCF      & 30                & 25.5             & 0.9725           & 0.9623            \\
JHU      & 75.25             & 30.5             & 0.9807           & 0.9532            \\
NWPU     & 86.15             & 35.0             & 0.9695           & 0.9486            \\ \hline
\end{tabular}%
}
\caption{Details of manual mask correction for datasets.}
\label{tab:manual mask correction for different datasets}
\vspace{-4mm}
\end{table}

\begin{figure*}[t]
  \centering
  \setlength{\abovecaptionskip}{-0.02cm} 
  \setlength{\belowcaptionskip}{-0.2cm} 
  \includegraphics[width=\linewidth]{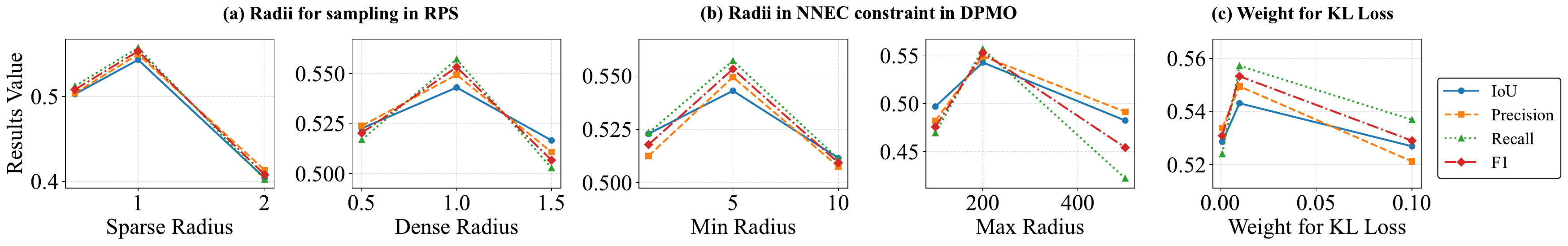}
  \caption{Results contrast for three hyperparameters in DPMO-RPS. (a) Different radii when sampling in RPS, (b) Different radii in the NNEC constraint of DPMO, (c) Different weights of the KL loss. 
}
  \label{fig:hyperparameters}
  \vspace{-0.2cm}
\end{figure*}

\vspace{-0.3cm}
\subsection{Implementation Details}
\label{subsec:imple}

All datasets were preprocessed by resizing images to ensure a minimum edge length of 512 pixels. The maximum edge length was set to 1536 for the UCF-QNRF~\cite{Idrees2018UCF-QNRF} dataset and 2048 for the JHU-Crowd++~\cite{sindagi2020jhu} and NWPU-Crowd~\cite{Wang2021NWPU_Crowd} datasets. 
The proposed DPMO-RPS framework was trained end-to-end with standard data augmentations, including random scaling, cropping, horizontal flipping, and color jittering. Optimization was performed using Adam with a learning rate of $1\times10^{-5}$ and a weight decay of $1\times10^{-4}$.

To improve annotation quality, we developed a web-based mask correction tool that allows annotators to refine the pseudo masks generated by DPMO directly on the original images. The average correction time and annotation quality on different datasets are reported in Tab.~\ref{tab:manual mask correction for different datasets}.

\begin{table}[t]
\centering
\setlength{\abovecaptionskip}{-0.01cm}
\setlength{\belowcaptionskip}{-0.5cm}
\resizebox{\columnwidth}{!}{%
\tiny
\begin{tabular}{ccccc}
\hline
model              & IoU $\uparrow$ & Precision $\uparrow$ & Recall $\uparrow$ & F1 $\uparrow$ \\ \hline
SAM+classification & 0.3254    & 0.2509          & 0.2437       & 0.2472  \\
FastSAM            & 0.3823    & 0.3125          & 0.2604       & 0.2841   \\
CrowdSAM           & 0.3048    & 0.2237          & 0.1098       & 0.1473   \\
\textbf{Ours}               & \textbf{0.5466}    & \textbf{0.5823}          & \textbf{0.5328}       & \textbf{0.5565}   \\ \hline
\end{tabular}%
}
\caption{Segmentation results on independent human annotated subset generated from crowd counting datasets.}
\label{tab:new dataset}
\end{table}

\vspace{-0.3cm}
\subsection{Experiments for Crowd Instance Segmentation}
\label{subsec:comparison for segmentation}

We compare our DPMO-RPS framework with three representative instance segmentation methods, namely SAM~\cite{kirillov2023segment}, FastSAM~\cite{zhao2023fast}, and CrowdSAM~\cite{cai2024crowd}, on the ShanghaiTech Part A~\cite{zhang2016single}, UCF-QNRF~\cite{Idrees2018UCF-QNRF}, JHU-Crowd++~\cite{sindagi2020jhu}, and NWPU-Crowd~\cite{Wang2021NWPU_Crowd} datasets. Quantitative results under the evaluation metrics described above are presented in Tab.~\ref{tab:Contrast with state-of-the-arts}. 
For SAM+Classification, segmentation masks were generated and filtered to retain only the "human" category. FastSAM and CrowdSAM were evaluated using their official checkpoints.

Qualitative comparisons are shown in Fig.~\ref{fig:Results Contrast}. Compared with SAM+Classification, the proposed framework more effectively separates foreground from background. Compared with FastSAM, it produces mask shapes that more accurately fit human heads. Furthermore, compared with CrowdSAM, our method achieves more accurate instance segmentation in densely crowded and heavily occluded regions.

\vspace{-0.3cm}
\subsection{Experiments for Crowd Counting}
\label{subsec:comparison for counting}
We list several classic crowd counting network models from recent years and select three models with the most accurate counting results for our experiments.
We evaluate our mask-supervised loss function on CLIP-EBC~\cite{ma2024clip} (density map prediction), PET~\cite{liu2023point} (point prediction), and ZIP~\cite{Ma2025ZIPSC} (density map prediction) models. As shown in Tab.~\ref{tab:loss function results}, under the supervision of our designed mask loss function, these crowd counting models significantly improve counting performance on the vast majority of datasets. 
Results indicate that the segmentation masks from our DPMO-RPS framework for crowd instance segmentation enhance counting accuracy in traditional crowd counting models.

\begin{table}[t] 
\centering
\setlength{\abovecaptionskip}{-0.01cm}
\setlength{\belowcaptionskip}{-0.05cm}
\resizebox{\columnwidth}{!}{%
\tiny
\begin{tabular}{clclclc}
\hline
 network  & & annotations  & &  MAE $\downarrow$   & &   RMSE  $\downarrow$      \\ \hline
 \multirow{3}{*}{ZIP\cite{Ma2025ZIPSC}}    & & points & & \underline{47.81} & & \underline{75.04} \\
            & & boxes   & & 48.68   & & 75.84     \\
             & & \textbf{masks} & & \textbf{46.95} & & \textbf{73.83}   \\ \hline
\end{tabular}%
}
\caption{Loss with different annotations on SHA dataset.}
\label{tab:Loss function supervised by boxes annotation}
\vspace{-0.3cm}
\end{table}

\begin{table}[t] 
\centering
\setlength{\abovecaptionskip}{-0.01cm}
\setlength{\belowcaptionskip}{-0.2cm}
\resizebox{\columnwidth}{!}{%
\begin{tabular}{ccccc}
\hline
          & IoU $\uparrow$    & Precision $\uparrow$ &  Recall $\uparrow$ &  F1  $\uparrow$    \\ \hline
w/o DPMO and RPS  & 0.0256  & 0.0133  &  0.0208 &  0.0162 \\
w/o DPMO   &  0.3867  &  0.3725    &  0.3778 &  0.3751  \\
w/o RPS    &  0.5131   &  0.5005    &  0.5081 &  0.5043 \\
\textbf{ours}  & \textbf{0.5431}  &  \textbf{0.5495}  &  \textbf{0.5572}  &  \textbf{0.5533} \\ \hline
\end{tabular}%
}
\caption{Segmentation results without different parts of DPMO-RPS framework on ShanghaiTech A dataset.}
\label{tab:ablation study}
\vspace{-0.3cm}
\end{table}

\begin{table}[t]
\centering
\setlength{\abovecaptionskip}{-0.01cm}
\setlength{\belowcaptionskip}{-0.2cm}
\resizebox{\columnwidth}{!}{%
\begin{tabular}{ccccc}
\hline
Method       & IoU  $\uparrow$  &  Precision $\uparrow$ & Recall $\uparrow$ &  F1 $\uparrow$  \\ \hline
GL~\cite{Wan2021GLloss} + SAM       & 0.1866  & 0.2241 & 0.0308 &  0.0542  \\
GL~\cite{Wan2021GLloss} + ours        &  \textbf{0.3187} & \textbf{0.3053 }   & \textbf{0.0849} &  \textbf{0.1328}  \\ \hline
CLTR~\cite{liang2022end} + SAM       & 0.2135    & 0.2069 & 0.0524 & 0.0836   \\
CLTR~\cite{liang2022end} + ours      & \textbf{0.3065} &  \textbf{0.2867}    & \textbf{0.1048} &  \textbf{0.1535}   \\ \hline
PET~\cite{liu2023point} + SAM       & 0.3024  &  0.2916 & 0.2831 &  0.2873   \\
\textbf{PET~\cite{liu2023point} + ours} & \textbf{0.4769} & \textbf{0.4426}  & \textbf{0.4396} & \textbf{0.4411}  \\ \hline
\end{tabular}%
}
\caption{Segmentation results with different point prediction models on UCF-QNRF dataset. }
\label{tab:results contrast with different nets}
\vspace{-4mm}
\end{table}

\vspace{-0.3cm}
\subsection{Ablation Studies}
\label{subsec:ablation}

\noindent\textbf{Results contrast on independent human annotated subset} 
To help address concerns that the evaluation protocol favors the proposed method, 
we randomly selected 80 images from all crowd datasets to form a new dataset and manually labeled the masks. The comparison results are shown in Tab.~\ref{tab:new dataset}.

\noindent\textbf{Effectiveness of mask supervision}
\quad To verify the performance of the proposed mask-supervised loss for crowd counting, we conducted experiments on the ShanghaiTech A~\cite{zhang2016single} dataset using the ZIP model~\cite{Ma2025ZIPSC}, as shown in Tab.~\ref{tab:Loss function supervised by boxes annotation}. 
We generate bounding box annotations by creating a bounding rectangle around the mask annotations in ShanghaiTech A dataset.
While bounding boxes offer richer spatial cues than point annotations, they inevitably capture background noise that can degrade model performance. Nevertheless, experimental results show that our method outperforms standard box-supervised losses on box-annotated datasets.

\noindent\textbf{DPMO}
\quad In our DPMO paradigm, the NNEC constraint significantly enhances segmentation capabilities of SAM. The experimental results on ShanghaiTech A~\cite{zhang2016single} dataset, shown in Tab.~\ref{tab:ablation study}, indicate that the absence of the NNEC constraint reduces segmentation accuracy. 

\noindent\textbf{RPS}
\quad We propose RPS trained with modified GRPO~\cite{shao2024deepseekmath} to select the best predicted point as a prompt to generate a better segmentation mask. 
We compared the difference in instance segmentation performance between models with and without RPS through experiments, as shown in Tab.~\ref{tab:ablation study}.
Results indicate that RPS can help further improve the accuracy of crowd instance segmentation.

\noindent\textbf{Point prediction network for RPS}
\quad As shown in Tab.~\ref{tab:results contrast with different nets}, integrating PET~\cite{liu2023point} into the DPMO-RPS framework achieves superior crowd segmentation compared to other point-based models, while substantially enhancing SAM's accuracy in dense crowd instance segmentation.

\vspace{-0.3cm}
\subsection{Hyperparameters}
\label{subsec:hyperparameters}

\noindent\textbf{Sampling radius in RPS}
\quad Initial point sampling aims to generate candidate prompts around the predicted head locations. 
Based on the results in Fig.~\ref{fig:hyperparameters}, the Gaussian radius is set to 1 for both sparse and dense regions.

\noindent\textbf{Radius in the NNEC constraint}
\quad The NNEC constraint uses a maximum enclosing circle to limit the mask size while ensuring that each point corresponds to a valid segmentation mask. According to the results in Fig.~\ref{fig:hyperparameters}, the minimum and maximum circle radii are set to 5 and 200, respectively.

\noindent\textbf{Weighting coefficient in modified GRPO loss}
\quad The KL divergence term in modified GRPO loss constrains the update magnitude of the policy, making its weighting coefficient an important hyperparameter. We evaluate three values (0.001, 0.01, and 0.1) and select 0.01, which achieves the best segmentation performance, as shown in Fig.~\ref{fig:hyperparameters}.

\vspace{-0.3cm}
\section{Conclusion}
\quad We propose a novel Dense Point-to-Mask Optimization with Reinforced Point Selection (DPMO-RPS) framework for crowd instance segmentation, including Dense Point-to-Mask Optimization (DPMO) with Nearest Neighbor Exclusive Circle (NNEC) constraint and Reinforced Point Selection (RPS) network trained with modified GRPO. 
The NNEC constraint in DPMO ensures non-overlapping masks with one-to-one correspondence to point annotations.
Then, the RPS predicts individual locations as point prompts for SAM in instance segmentation.
End-to-end training of our DPMO-RPS framework, we achieve the state-of-the-art crowd instance segmentation performance on ShanghaiTech, UCF-QNRF, JHU-Crowd++, and NWPU-Crowd datasets. 
Furthermore, we design two loss functions based on density map prediction and point prediction, which are supervised by robust mask annotations with manual corrections. The improvement of counting accuracy across multiple models indicates that segmentation masks also play an important role in improving prediction and counting accuracy.

\bibliography{aaai2027}


\end{document}